\documentclass[11pt]{article}

\usepackage[margin=1in]{geometry}
\usepackage[T1]{fontenc}
\usepackage{times}
\usepackage{amsmath,amssymb,amsthm}
\usepackage{graphicx}
\usepackage{booktabs}
\usepackage{xcolor}
\usepackage{url}
\usepackage[round]{natbib}
\usepackage{caption}
\usepackage{subcaption}
\usepackage[colorlinks=true,linkcolor=blue!60!black,citecolor=blue!60!black,urlcolor=blue!60!black]{hyperref}

\makeatletter
\renewcommand{\p@subfigure}{\thefigure}
\makeatother

\newtheorem{theorem}{Theorem}
\newtheorem{proposition}[theorem]{Proposition}
\newtheorem{lemma}[theorem]{Lemma}
\newtheorem{corollary}[theorem]{Corollary}
\newtheorem{assumption}{Assumption}

\newsavebox{\figabox}
\newcommand{\projectsite}{\href{https://srikarbabug.github.io/ff-safemarl}{\texttt{srikarbabug.github.io/ff-safemarl}}}

\title{Distilling Safety Filters for Filter-free Multi-agent RL}
\author{%
  Srikar Babu Gadipudi \qquad Somayeh Sojoudi \qquad Claire J. Tomlin\\[10pt]
  University of California, Berkeley\\[10pt]
  \texttt{\{srikarbabu, sojoudi, tomlin\}@berkeley.edu}
}
\date{}

\begin{document}

\maketitle

\begin{abstract}%
\noindent Reinforcement learning (RL) can teach robot teams to navigate shared spaces, but trial and error causes collisions that real robots cannot afford. Control barrier function (CBF) safety filters prevent these collisions by correcting unsafe actions, yet in a team the filter becomes a centralized optimization whose number of constraints grows quadratically with the team size. We propose an approach that builds the filter's safety into the RL policy and examine how such a system behaves during filter-free deployment. However, the filter that keeps training safe also hides what the policy must learn: it masks the cost of unsafe actions, which actor-critic methods then overvalue, and it keeps near-collision states out of the training data. Our approach therefore learns only from executed safe actions with implicit Q-learning and distills the filter into the policy at synthetic near-collision states. We prove that filter distillation loss bounds the policy's constraint violation and hence how far agents can cross their safety margin, so errors never accumulate. When distillation states cover deployed states, the loss also bounds the collision probability at every step. Without a filter and acting only on local observations, our policies have no collisions in $1{,}800$ test episodes with multiple agents. A policy trained with three agents transfers zero-shot to teams of twenty at constant per-agent computation and stays collision-free with up to eight agents. We test our policies on Crazyflie quadrotors with up to five flying at the same time.\footnote{Videos of the hardware experiments and code are available at \projectsite.}
\end{abstract}

% \noindent\textbf{Keywords:} Safe multi-agent reinforcement learning, Control barrier functions, Policy distillation, Zero-shot decentralized collision avoidance

\section{Introduction}
\label{sec:intro}

Robots increasingly work in teams that share space, where each robot must reach its goal while avoiding collisions with other robots and obstacles. Reinforcement learning (RL) offers a promising approach to learning such behaviours, but its trial-and-error nature causes collisions during training that can damage hardware, so real-world multi-agent RL requires safe exploration from the start. Control barrier function (CBF) safety filters \citep{ames2017, ames2019} address this challenge: at each step, the filter minimally modifies the actions proposed by the policy to satisfy the collision-avoidance constraints, so only safe actions are executed during training \citep{fisac2018general}. In multi-agent systems, however, pairwise constraints couple the team into a centralized quadratic program (QP) whose number of constraints grows quadratically with the team size. Solving it requires every agent's state and proposed action; for $20$ agents, the joint QP takes $16.9$\,ms, a third of the control period, against $0.6$\,ms per agent for our policy (Sec.~\ref{sec:scaling}). This motivates our objective: \emph{use a centralized safety filter to guarantee collision-free exploration during training, but learn a decentralized policy that no longer requires the filter at deployment.}

Simply removing the filter after training, however, leads to a fundamental distribution mismatch: the policy is trained under filtered action-to-transition mapping but deployed under an unfiltered one. CBF-RL \citep{yang2025} adds barrier-based reward shaping so that single-agent policies, trained with the filter, run without it at deployment. Because the filter maps many raw actions to the same safe action \citep{markgraf2025}, actor-critic methods such as soft actor-critic (SAC) can overvalue unsafe raw actions: SAC trained with the filter in the loop collides in $98\%$ of episodes when deployed without it (Sec.~\ref{sec:ablations}).

This observation motivates RL methods that learn only from actions in the data: we extract the policy from the executed safe actions with implicit Q-learning (IQL) \citep{kostrikov2022} (Sec.~\ref{sec:learn-safe}). IQL alone is insufficient, however: because the filter prevents collisions, the replay buffer lacks the near-collision states that the unfiltered policy may reach. To cover them, off-support safety-filter distillation (OSFD; Sec.~\ref{sec:osfd}) trains the policy on the actions of the training filter, queried at synthetic near-collision states, without environment interaction. Following centralized training with decentralized execution, only training uses the joint state, and the deployed policy is an attention-based actor on local observations whose parameters do not depend on the team size (Sec.~\ref{sec:po}).

We prove theoretically that this training controls safety once the filter is removed (Sec.~\ref{sec:theory}). The distillation loss bounds a closed-form CBF residual that measures how much the unfiltered policy violates the filter's constraints, a bounded residual lowers the barrier by a bounded amount uniformly in time, so violations never accumulate, and, when the distillation states cover the deployed states, the loss bounds the collision probability at every step. The training loss thus certifies the safety of the filter-free policy, and every bound held on the $75$ trained policies we audited. 
% In safety-critical applications, independent safeguards such as a separately designed safety filter would still be used at deployment, but a policy with safety built in should rarely need them.

Our contributions are summarized as follows: (1) Filter-free multi-agent RL: we propose a novel framework that trains collision-free with a centralized CBF filter and deploys a decentralized policy without it, by combining IQL-based policy extraction with OSFD. (2) Safety guarantees: bounds that tie the distillation loss to the safety of the unfiltered policy under single- and double-integrator dynamics (Sec.~\ref{sec:theory}). (3) Scalable filter-free collision avoidance: our local-sensing policies have no collision in $1{,}800$ episodes on the training scenes, where CBF-RL collides in up to $44\%$, a policy trained with three agents transfers zero-shot to twenty at a constant $0.6$\,ms per agent (Sec.~\ref{sec:exp}), and our policies fly up to five Crazyflie quadrotors without a filter (Sec.~\ref{sec:hardware}).

\section{Related Work}
\label{sec:related-work}

\paragraph{Safe RL with safety filters.}
Constrained RL \citep{achiam2017} bounds expected cumulative costs but not state-wise safety during training. Safety filters enforce constraints at every step through shielding \citep{alshiekh2018}, safety layers \citep{dalal2018}, Hamilton-Jacobi reachability \citep{fisac2018general, bansal2017}, also learned by RL \citep{fisac2019}, CBFs \citep{cheng2019, bejarano2024} or control barrier-value functions that unify the two \citep{choi2021cbvf}. Learned filters handle model uncertainty \citep{choi2020} and black-box dynamics \citep{lavanakul2024}, but generally remain in loop at execution. CBF-RL \citep{yang2025} trains with a CBF filter and shapes reward so the policy runs without it. We instead address coupled multi-agent constraints and decentralized policies under partial observability. We close the train-deploy mismatch by in-support policy extraction and supervision beyond data rather than reward shaping.

\paragraph{Multi-agent collision avoidance and safety constraints.}
Learning-based collision avoidance has been explored for ground and aerial swarms \citep{chen2017cadrl, long2018, batra2021, huang2024}. Decentralized barrier certificates \citep{wang2017}, shields \citep{elsayed2021, melcer2022} and CBF filters within multi-agent RL \citep{cai2021} avoid a central solve by splitting the constraints among agents, but remain in the loop at deployment, and neural approaches learn decentralized CBFs together with their controllers \citep{qin2021, zhang2023gcbf, zhang2025gcbf+}. In centralized training with decentralized execution (CTDE) \citep{lowe2017, yu2022mappo}, critics use joint information during training, while each deployed policy acts on its own observation. We follow CTDE and additionally use a centralized filter for safe exploration and as a teacher for a filter-free decentralized policy.

\paragraph{Data-driven RL, imitation learning and partial observability.}
Offline RL counters value overestimation at actions absent from the data \citep{levine2020} by constraining the policy to the behavior policy \citep{fujimoto2019, kumar2019bear, wu2019brac, nair2020, fujimoto2021td3bc}, by penalizing out-of-distribution values \citep{kumar2020} or by learning values in-sample \citep{kostrikov2022, xu2023ivr, garg2023xql}, also with multiple agents \citep{yang2021icq, pan2022omar}. In-sample learning suits filtered transitions, where unsafe raw actions share safe outcomes, but leaves states missing from the buffer unsupervised. DAgger \citep{ross2011} queries an expert on the learner's own roll-outs; OSFD instead queries the training filter at synthetic near-collision states, without environment interaction. As in learning with privileged information \citep{chen2020lbc, lee2020}, we use the global state only during training and our local-observation attention actor builds on \citet{zhang2025gcbf+}.

\section{Our Approach}
\label{sec:approach}

\begin{figure}[t]
    \centering
    \sbox{\figabox}{\includegraphics[width=0.72\linewidth]{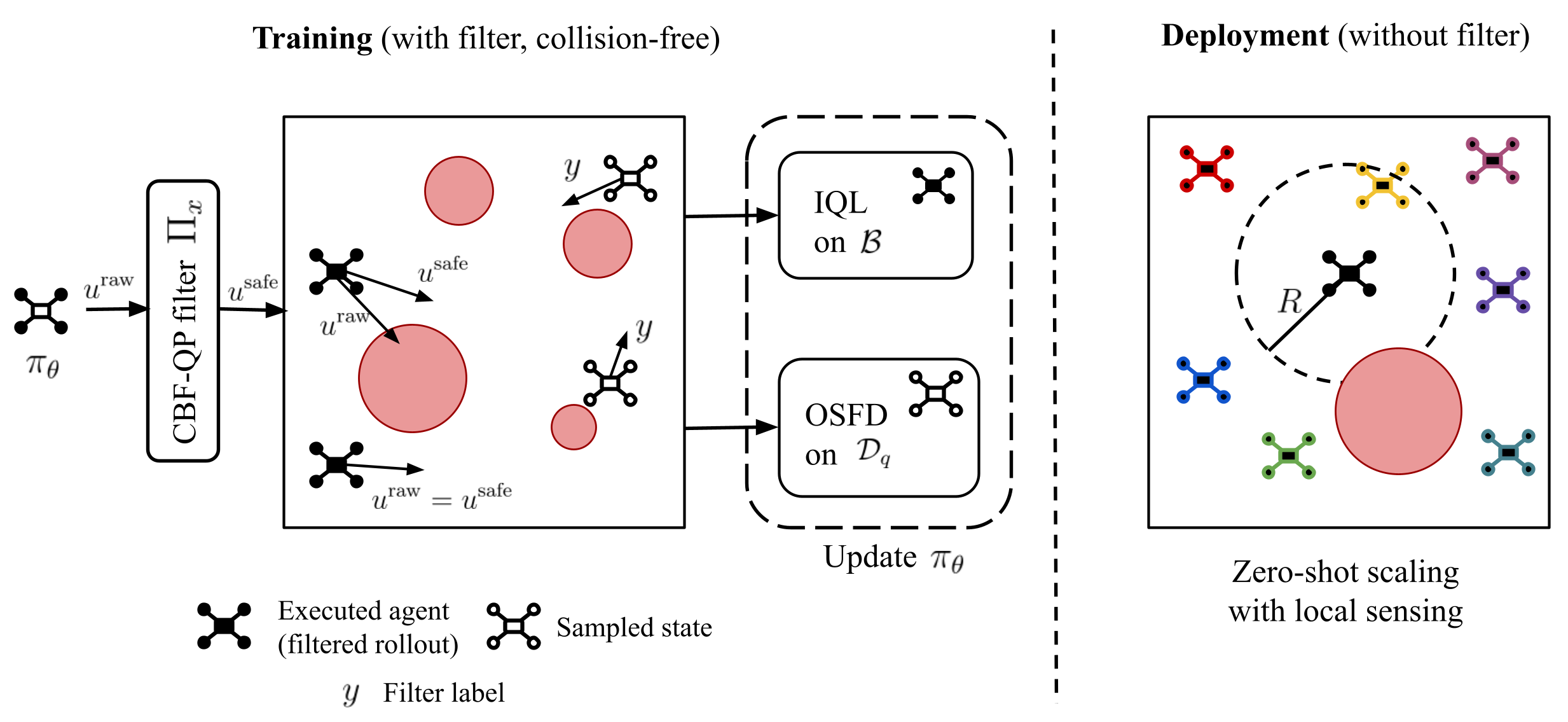}}
    \begin{subfigure}[b]{0.72\linewidth}
        \usebox{\figabox}
        \caption{Training and deployment}\label{fig:method}
    \end{subfigure}\hfill
    \begin{subfigure}[b]{0.23\linewidth}
        \centering
        \raisebox{0.5\dimexpr\ht\figabox+\dp\figabox-\height-\depth\relax}{%
          \setlength{\fboxsep}{0pt}\setlength{\fboxrule}{0.8pt}%
          \fbox{\includegraphics[width=\dimexpr\linewidth-1.6pt\relax]{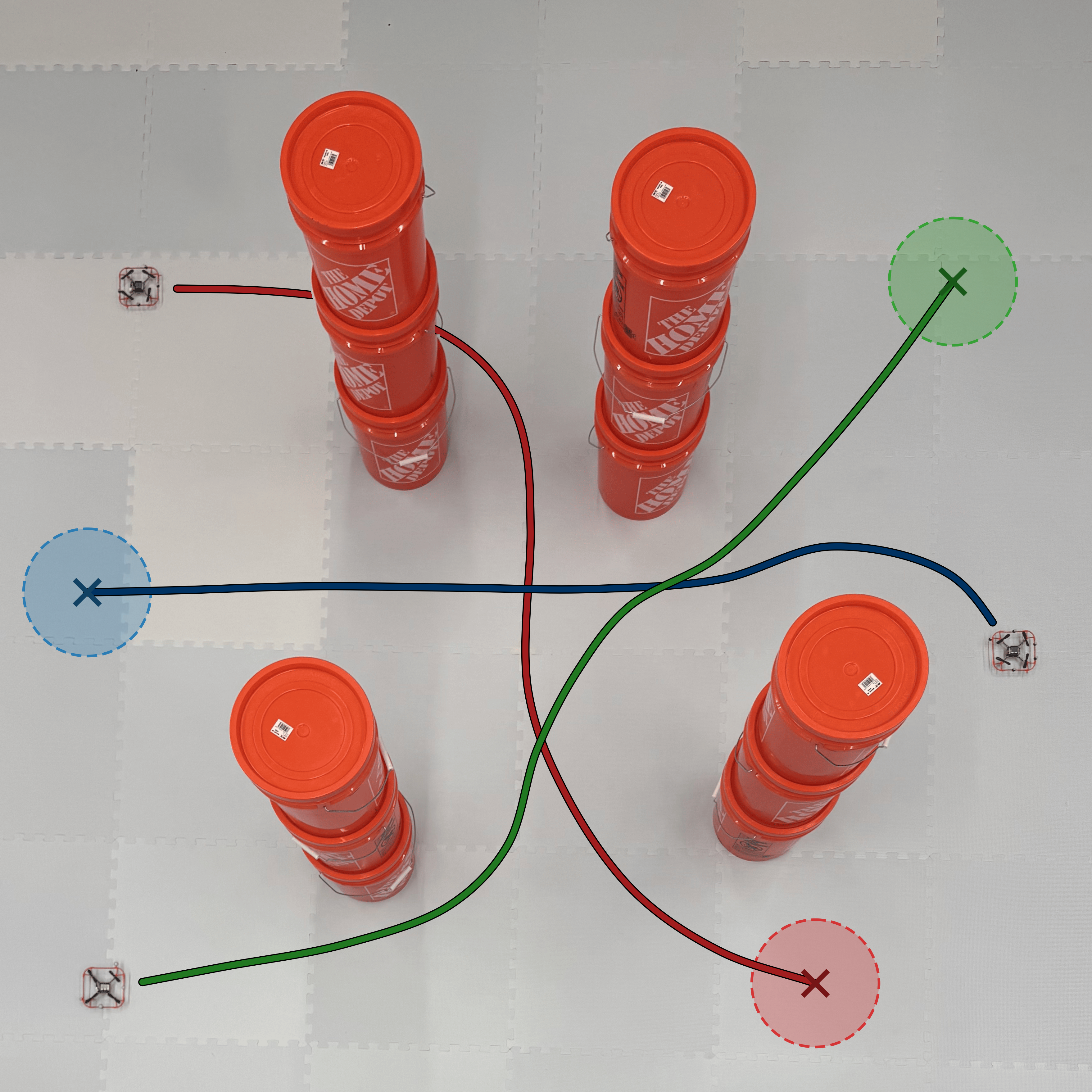}}}
        \caption{Hardware arena}\label{fig:hardware}
    \end{subfigure}
    {\caption{(a) The CBF-QP filter $\Pi_x$ keeps training collision-free, the policy learns with IQL on the filtered buffer $\mathcal{B}$ and with OSFD on filter labels $y$ at sampled states and is deployed without the filter, with local sensing within $R$. (b) Three Crazyflies and four stacked-bucket obstacles (Sec.~\ref{sec:hardware}), lines are illustrative paths to the goals.}}
    \label{fig:overview}
\end{figure}

We now present our approach for learning decentralized collision-avoidance policies using a centralized safety filter during training (Fig.~\ref{fig:method}). The actor is trained with two losses: an RL loss on the executed safe actions (Sec.~\ref{sec:learn-safe}) and a distillation loss at synthetic near-collision states (Sec.~\ref{sec:osfd}).

% \begin{figure}
%     \centering
%     \includegraphics[width=\linewidth]{figures/safemarl-overview.png}
%     \caption{Training (left): the CBF-QP filter $\Pi_x$ keeps training collision-free, and the policy learns with IQL on the filtered buffer $\mathcal{B}$ and with OSFD on filter labels $y$ at sampled states. Deployment (right): no filter, only local sensing with radius $R$.}
%     \label{fig:overview}
% \end{figure}

\subsection{Problem Formulation}
\label{sec:formulation}

We consider $n$ disc-shaped agents of radius $r$ navigating toward individual goals $g_i$ in an environment with circular obstacles. Each agent follows either single-integrator (SI) or double-integrator (DI) dynamics, discretized with time step $\Delta t$:
\begin{equation}
\label{eq:dynamics}
\mathrm{SI}:\ p_{i,t+1} = p_{i,t} + \Delta t\, u_{i,t},\qquad
\mathrm{DI}:\ v_{i,t+1} = v_{i,t} + \Delta t\, u_{i,t}, \quad p_{i,t+1} = p_{i,t} + \Delta t\, v_{i,t+1}.
\end{equation}
Here, $p_i$, $v_i$ and $u_i$ are position, velocity and control, with $u\in\mathcal{U}=\{u:\|u_i\|_\infty\le u_{\max}\ \forall i\}$ and, for DI agents, $\|v_i\|_\infty\le v_{\max}$. Unsubscripted norms are Euclidean. Agent $i$ observes $o_i$, either the full scene or its own state and the agents and obstacles within a sensing radius $R$, and receives an individual reward for goal progress, goal completion and collisions (App.~\ref{app:impl}). We seek a policy shared by all agents, a Gaussian $\pi_\theta$ squashed by $u_{\max}\tanh\mu_\theta(o_i)$, that is trained without collisions and deployed on local observations without a filter or communication.

\paragraph{Safety filter.} During training, a centralized CBF filter corrects unsafe actions. With $x$ joint state, we index by $c$ agent-obstacle, -wall and -agent constraints and give each constraint a barrier $h_c$:
\begin{equation}
\label{eq:barriers}
\begin{aligned}
h_{i,o} = \|p_i - c_o\| - (r_o + r + \rho),\quad
h_{i,\mathrm{wall}} = (L - r - \rho) \mp (p_i)_k,\quad
h_{ij} = \|p_i - p_j\| - (2r + \rho_{\mathrm{pair}}),
\end{aligned}
\end{equation}
where obstacle $o$ has center $c_o$ and radius $r_o$, $L$ is arena half-width and $\rho,\rho_{\mathrm{pair}}>0$ are margins. With margin $\rho_c$ for $c$, tightened safe set is $\mathcal{S}=\{x:h_c(x)\ge0\ \forall c\}$; collision requires $h_c(x)<-\rho_c$. A CBF makes $\mathcal{S}$ forward invariant: if the CBF condition $\dot h_c+\alpha h_c\ge0$ holds along the trajectory, $h_c$ never becomes negative, so a system that starts in $\mathcal{S}$ stays in it \citep{ames2017}. This condition for SI dynamics and the high-order CBF (HOCBF) condition $\dot\psi_{1,c}+\alpha_2\psi_{1,c}\ge0$ with $\psi_{1,c}=\dot h_c+\alpha_1h_c$ \citep{xiao2019} for DI dynamics each give an affine constraint $a_c(x)^T u\ge b_c(x)$ on the joint control, the row of $c$ in the filter below, with $a_c=\nabla h_c$ and $b_c=-\alpha h_c$ under SI dynamics (App.~\ref{app:filter}).

Given the joint action $u^{\mathrm{raw}}$ proposed by the agents, the filter solves \citep{wang2017}
\begin{equation}
\label{eq:filter}
\begin{aligned}
\Pi_x(u^{\mathrm{raw}}) = \operatorname*{arg\,min}_{u \in \mathcal{U},\, s \geq 0}\quad & \tfrac{1}{2}\|u - u^{\mathrm{raw}}\|^2 + \tfrac{\kappa}{2}\|s\|^2\\
\text{s.t.}\quad & a_c(x)^T u + \sigma s_c \geq b_c(x) \quad \forall c.
\end{aligned}
\end{equation}
Only $u^{\mathrm{safe}}=\Pi_x(u^{\mathrm{raw}})$ is executed. Under SI dynamics the constraints are hard ($\sigma=0$), feasible on $\mathcal{S}$, and keep the system in $\mathcal{S}$ (Prop.~\ref{prop:si-train}). Under DI dynamics, states with $\psi_{1,c}<0$, such as an agent coasting toward an obstacle, are reachable and may admit no feasible acceleration \citep{xiao2019}, so a penalized slack ($\sigma=1$) keeps the QP feasible (App.~\ref{app:filter}). The transitions $(o,u^{\mathrm{raw}},u^{\mathrm{safe}},r,o')$, with $o=(o_1,\dots,o_n)$ and $o'$ as current and next-step observations and rewards $r=(r_1,\dots,r_n)$, are stored in a replay buffer $\mathcal{B}$.

\subsection{Learning from Safe Experience}
\label{sec:learn-safe}

Under the filter, many raw actions map to the same executed action and to identical transitions. This action aliasing \citep{markgraf2025} lets actor-critic methods overvalue unsafe proposals whose consequences the filter corrected, so we extract the policy from the executed safe actions.

We adopt IQL \citep{kostrikov2022} with centralized critics and per-agent values \citep{lowe2017}: with $Q^i_{\phi_k}$ $k$-th critic of agent $i$, $V^i_\psi$ its value and $\bar\phi_k$ parameters of target critic, we minimize
\begin{equation}
\label{eq:iql}
\begin{aligned}
\mathcal{L}_V = \sum_i \mathbb{E}_{\mathcal{B}}\Big[\ell_\tau\Big(\min_k Q^i_{\bar{\phi}_k}(o, u^{\mathrm{safe}}) - V^i_\psi(o)\Big)\Big],\quad
\mathcal{L}_Q = \sum_{i,k} \mathbb{E}_{\mathcal{B}}\Big[\Big(Q^i_{\phi_k}(o, u^{\mathrm{safe}}) - r_i - \gamma V^i_\psi(o')\Big)^2\Big],
\end{aligned}
\end{equation}
where $\ell_\tau(z) = |\tau - \mathbf{1}[z < 0]|\, z^2$ is the expectile regression loss. The policy is then extracted through advantage-weighted regression \citep{peng2019},
\begin{equation}
\label{eq:awr}
\mathcal{L}_{\mathrm{AWR}} = \mathbb{E}_{\mathcal{B}}\bigg[\sum_{i=1}^{n} W_i \Big\|\tanh \mu_\theta(o_i) - \frac{u^{\mathrm{safe}}_i}{u_{\max}}\Big\|^2\bigg], \qquad W_i \propto \exp(\beta A_i),\quad A_i = Q^i_{\phi}(o, u^{\mathrm{safe}}) - V^i_\psi(o),
\end{equation}
with temperature $\beta>0$. Minimizing it regresses the policy onto the executed safe actions, weighted toward higher advantage (weights are clipped, normalized and up-weighted where the filter intervened; App.~\ref{app:learning}). Since every regression target satisfies the filter's constraints, the population optimum satisfies every obstacle, wall and agent-agent constraint the agent senses.

% the extracted policy's residual on the training data is zero at the optimum (\red{add proposition reference}).

\paragraph{Why IQL?} Unlike SAC, IQL never maximizes the critic over arbitrary raw actions. Most offline RL methods \citep{fujimoto2019, kumar2020, fujimoto2021td3bc} still bootstrap in the next actions sampled from the policy, each of which would need a joint filter to follow the executed dynamics (App.~\ref{app:iql}). IQL instead bootstraps from $V_\psi(o')$ and evaluates only buffered actions.

\subsection{Off-support Safety-Filter Distillation}
\label{sec:osfd}

To supervise the policy beyond the replay buffer, OSFD queries the training filter at synthetic safety-critical states instead of executing additional actions. At each update, we draw joint states from a distribution $\mathcal{D}_q$ that mixes buffer states, perturbed buffer states and synthetic configurations $\mathcal{D}_{\mathrm{off}}$ that place agents close to obstacles, walls or one another, moving toward them. Under DI dynamics, these include states with $\psi_{1,c}<0$, which the filtered trajectories rarely visit. Each state is labeled by the training filter, $y(x)=\Pi_x\big(u^{\mathrm{ref}}(x)\big)$, where $u^{\mathrm{ref}}$ is a goal-directed reference controller with a tangential component against deadlock \citep{reis2021}. The actor imitates these labels with
\begin{equation}
\label{eq:osfd}
\mathcal{L}_{\mathrm{OSFD}} = \mathbb{E}_{x \sim \mathcal{D}_q}\bigg[\frac{1}{|\mathcal{A}|} \sum_{i \in \mathcal{A}} \Big\|\tanh \mu_\theta(o_i) - \frac{y_i(x)}{u_{\max}}\Big\|^2\bigg],
\end{equation}
where $\mathcal{A}$ excludes agents held at their goals (``active agents'').

OSFD needs no environment interaction and, unlike DAgger \citep{ross2011}, never executes the learner's action, so training remains collision-free. A penalty on the part of the policy's action that violates the filter's constraints further sharpens the policy at the constraint boundary (App.~\ref{app:learning}).

\subsection{Partial Observability and Decentralized Execution}
\label{sec:po}

We follow CTDE \citep{lowe2017} with a strict split: the filter, critics and OSFD labels use the joint state $x$ only during training, and the deployed system is the actor alone, acting on local observations. Agent $i$ senses the set $\mathcal{N}_i(x)$ of agents whose center, and obstacles whose surface, lies within $R$ of $p_i$. It observes an ego vector $e_i$ (goal direction and distance, wall clearances, previous action and, under DI dynamics, its velocity) and one token $\tau_{ik}=(n_{ik},h_{ik},r_k,v_k-v_i,\chi_k)$ per sensed entity $k$. Each token holds the normal $n_{ik}=\nabla_{p_i}h_{ik}$ and value of the filter's barrier \eqref{eq:barriers} between $i$ and $k$, the radius, the relative velocity and the type $\chi_k$ (one-hot: agent or obstacle), so the actor perceives safety in the filter's coordinates. The ego features query the encoded tokens $\phi_\theta(\tau_{ik})$ by cross-attention \citep{vaswani2017},
\begin{equation}
\label{eq:attention}
\mu_\theta(o_i)=f_\theta\Big(e_i,\ \mathrm{CrossAttn}\big(e_i,\ \{\phi_\theta(\tau_{ik})\}_{k\in\mathcal{N}_i\cup\{\varnothing\}}\big)\Big),
\end{equation}
where $\varnothing$ is a learned empty token and $f_\theta$ is an MLP. Attention aggregates any number of tokens into one fixed-size vector, so the actor's parameters depend on neither the number of agents nor that of obstacles, and a policy trained with three agents runs \emph{unchanged} on larger teams. Each agent's computation grows with $|\mathcal{N}_i|$ and needs no communication.

The actor minimizes $\beta_d\mathcal{L}_{\mathrm{AWR}}+w\mathcal{L}_{\mathrm{OSFD}}$, where the weights $\beta_d$ and $w$ balance reward against imitation of the filter (Sec.~\ref{sec:ablations}), with the constraint-violation penalty and an entropy term (App.~\ref{app:learning}). Gradients never pass through the QP. At deployment, each agent executes $u_i=u_{\max}\tanh\mu_\theta(o_i)$.

\section{Deployment-Time Safety Analysis}\label{sec:theory}

The deployed system is the closed loop \eqref{eq:dynamics} under the joint action $\pi(x)=(\pi_\theta(o_1),\dots,\pi_\theta(o_n))$ without a filter. For the filter row $a_c(x)^T u\ge b_c(x)$ of a constraint $c$ in \eqref{eq:filter} and the OSFD label $y(x)$, we define the policy's \emph{CBF residual}, its \emph{label error} and the label's own violation,
\begin{equation}
\label{eq:residual}
\nu_c(x)=\big[b_c(x)-a_c(x)^T\pi(x)\big]_+,\qquad
e_\theta(x)=\big\|\pi(x)-y(x)\big\|,\qquad
\epsilon_c(x)=\big[b_c(x)-a_c(x)^T y(x)\big]_+,
\end{equation}
where $[z]_+=\max(z,0)$. The residual is closed-form and needs no QP. OSFD minimizes the label error, $\mathbb{E}_{\mathcal{D}_q}[e_\theta^2]\le n u_{\max}^2\mathcal{L}_{\mathrm{OSFD}}$ when every agent active. The label violation vanishes wherever the hard rows are feasible, in particular on $\mathcal{S}$ under SI dynamics, and is at most the slack under DI dynamics (App.~\ref{app:filter}). Our guarantee chains three bounds,
\begin{equation}
\label{eq:chain}
\underbrace{\mathcal{L}_{\mathrm{OSFD}}}_{\text{training loss}}
\ \xrightarrow{\ \eqref{eq:osfd}\ }\
\underbrace{e_\theta}_{\text{label error}}
\ \xrightarrow{\ \text{Prop.~\ref{prop:lossres}}\ }\
\underbrace{\nu_c}_{\text{CBF residual}}
\ \xrightarrow{\ \text{Thm.~\ref{thm:cert}}\ }\
\underbrace{h_c}_{\text{barrier}},
\end{equation}
and in the following, Corollary~\ref{cor:risk} turns the last bound into a collision probability.

\begin{proposition}[Distillation bounds the residual]\label{prop:lossres}
    For every labeled state $x$ and every constraint $c$, $\nu_c(x)\le\|a_c\|e_\theta(x)+\epsilon_c(x)$. Consequently, with every agent active,
    \begin{equation}
\label{eq:lossbound}
\Big(\mathbb{E}_{\mathcal{D}_q}\big[\nu_c^2\big]\Big)^{1/2}\ \le\ \|a_c\|\,u_{\max}\sqrt{n\,\mathcal{L}_{\mathrm{OSFD}}}\ +\ \Big(\mathbb{E}_{\mathcal{D}_q}\big[\epsilon_c^2\big]\Big)^{1/2},
\end{equation}
where $\|a_c\|=\|\nabla h_c\|$ equals $1$ for obstacle and wall constraints and $\sqrt{2}$ for pair constraints, under both dynamics.
\end{proposition}
\begin{proof}
    $b_c - a_c^T\pi=(b_c-a_c^T y)+a_c^T(y-\pi)\le\epsilon_c+\|a_c\|e_\theta$ by Cauchy--Schwarz, and the right side is non-negative. Minkowski's inequality and $\mathbb{E}_{\mathcal{D}_q}[e_\theta^2]\le n\,u_{\max}^2\mathcal{L}_{\mathrm{OSFD}}$ give \eqref{eq:lossbound}.
\end{proof}

Proposition~\ref{prop:lossres} uses only the label's feasibility, so it holds for any reference controller, both dynamics and any observation model. Under SI dynamics, $a_c^T\pi=\dot h_c$, so $\nu_c$ is the rate at which the policy decreases $h_c$ beyond the rate $\alpha h_c$ that the CBF condition allows.

\begin{theorem}[The residual bounds the barrier]\label{thm:cert}
Let $\alpha>0$ be the CBF gain, $\Delta t$ the time step with $a=\alpha \Delta t \in (0,1)$, and $\rho_c$ the margin of constraint $c$. Along the unfiltered SI loop, for every constraint $c$ and every $t\ge0$,
\begin{equation}
h_c(x_t)\ \ge\ (1-a)^t\,h_c(x_0)\ -\ \Delta t\sum_{k=0}^{t-1}(1-a)^k\,\nu_c(x_{t-1-k}) .
\label{eq:recursion}
\end{equation}
Consequently, for $x_0\in\mathcal{S}$: (i) if $\nu_c(x_t)\le \bar{\nu}$ for all $t$, then $h_c(x_t)\ge -\bar{\nu}/\alpha$ for all $t$, for any horizon; (ii) if $\nu_c(x_t) < \alpha \rho_c$ for all $t<T$, there is no collision on $c$ during the first $T$ steps.
\end{theorem}
\begin{proof}[Proof sketch]
    Convexity of the barriers (Lemma~\ref{lem:convex}) and \eqref{eq:residual} give one-step bound $h_c(x_{t+1})\ge(1-a)\,h_c(x_t)-\Delta t\,\nu_c(x_t)$. Unrolling it gives \eqref{eq:recursion}, and summing the geometric series gives (i) and (ii) (full proof in App.~\ref{app:proofs}).
\end{proof}

Residual errors therefore do not accumulate: \eqref{eq:recursion} contracts at rate $1-\alpha\Delta t$, so the barrier falls below $-\rho_c$ only after a sustained run of large residuals. By convexity of the barriers, the bound is exact in discrete time and needs no Lipschitz constant of the policy or the dynamics. To bound the collision probability, the deployed states must also be represented in $\mathcal{D}_q$.

\begin{assumption}[Coverage]\label{ass:coverage}
    There is a constant $C\ge 1$ such that, for every $t\le T$, the distribution $d_t$ of the deployed state $x_t$ satisfies $d_t(E)\le C \mathcal{D}_q(E)$ for every set of states $E$.
\end{assumption}

$C$ is a concentrability coefficient, standard in offline RL \citep{munos2003, chen2019}. It bounds how much more often the deployed policy visits any region than OSFD samples it, and, as in offline RL, coverage of the deployed policy alone suffices \citep{rashidinejad2021}. Unlike a fixed dataset, $\mathcal{D}_q$ is ours to design, and since the proof uses $C$ only through $\mathbb{E}_{d_t}[\nu_c^2]\le C\,\mathbb{E}_{\mathcal{D}_q}[\nu_c^2]$, $C$ can be measured (App.~\ref{app:audit}).

\begin{corollary}[Collision probability]\label{cor:risk}
    Under Assumption~\ref{ass:coverage} and SI dynamics, with $x_0\in \mathcal{S}$, for every constraint $c$ and every step $t\le T$,
    \begin{equation}
    \label{eq:prob}
    \Pr\big(h_c(x_t)<-\rho_c\big)\ \le\ \frac{C\,\mathbb{E}_{\mathcal{D}_q}\big[\nu_c^2\big]}{(\alpha\rho_c)^2}\ \le\ \frac{C}{(\alpha\rho_c)^2}\Big(\|a_c\|\,u_{\max}\sqrt{n\,\mathcal{L}_{\mathrm{OSFD}}}+\big(\mathbb{E}_{\mathcal{D}_q}\big[\epsilon_c^2\big]\big)^{1/2}\Big)^2 .
    \end{equation}
    Consequently, the probability of a collision on $c$ within $T$ steps is at most $T$ times the right side.
\end{corollary}
\begin{proof}[Proof sketch]
    By \eqref{eq:recursion}, a collision at step $t$ requires the discounted residual sum $\Delta t\sum_k(1-\alpha\Delta t)^k\nu_c(x_{t-1-k})$ to exceed $\rho_c$. By Minkowski's inequality and Assumption~\ref{ass:coverage}, its second moment is at most $C\,\mathbb{E}_{\mathcal{D}_q}[\nu_c^2]/\alpha^2$. Markov's inequality gives the first bound, Proposition~\ref{prop:lossres} the second, and a union bound over the $T$ steps the last claim (App.~\ref{app:proofs}).
\end{proof}

Corollary~\ref{cor:risk} turns the training loss into a deployment guarantee: every term on its right side is set during training, and the bound holds at every step for any horizon. It also shows why filtered data fall short: the unfiltered loop departs from the filtered one only where the residual is positive (Lemma~\ref{lem:noshift}), reaching states absent from the buffer that only OSFD covers. On $75$ trained policies, every link of \eqref{eq:chain} held, and a smaller loss went with a smaller residual and fewer collisions (App.~\ref{app:audit}).

\paragraph{Local observation and double integrators.} These results hold for any observation model. A decentralized policy cannot match the pair labels, which depend on the neighbor's intent (Prop.~\ref{prop:labelvar}), but if every agent keeps the violation of its own half of each pair row below $\frac12\alpha\rho_{\mathrm{pair}}$, no two agents collide (Prop.~\ref{prop:pair}). Under DI dynamics, Proposition~\ref{prop:lossres} applies unchanged, and two nested contractions bound the barrier uniformly in time up to an $O(\Delta t)$ discretization term (Theorem~\ref{thm:di}, App.~\ref{app:di}). This term can use most of the margin at our control step, so we measure DI safety empirically.

\section{Experiments}\label{sec:exp}

\begin{table}[t]
\centering\small
\caption{Unfiltered deployment on the training scenes: collisions ($\%$ of episodes) / goals reached ($\%$ of agents), 3 seeds $\times$ 100 episodes in simulation.}
\label{tab:main}
\begin{tabular}{llcccc}
\toprule
& & \multicolumn{2}{c}{Single integrator} & \multicolumn{2}{c}{Double integrator}\\
\cmidrule(lr){3-4}\cmidrule(lr){5-6}
Agents / obstacles & Obs. & Ours & CBF-RL & Ours & CBF-RL\\
\midrule
1 / 4 & full  & 0.3 / 99.3 & 0.3 / 99.3 & \textbf{0.0} / \textbf{100} & 4.0 / 97.3\\
      & local & \textbf{0.0} / \textbf{100}  & 0.0 / 99.3 & \textbf{0.0} / \textbf{100} & 2.7 / 99.0\\
3 / 2 & full  & \textbf{0.0} / \textbf{100} & 3.3 / 94.3 & 0.3 / 100 & 23.7 / 100\\
      & local & \textbf{0.0} / \textbf{100} & 5.7 / 37.7 & \textbf{0.0} / \textbf{100} & 22.7 / 99.3\\
3 / 4 & full  & 7.3 / 99.7 & 20.7 / 0.2 & 8.3 / 98.9 & 67.3 / 12.4\\
      & local & \textbf{0.0} / \textbf{95.4} & 4.7 / 0.0 & \textbf{0.0} / \textbf{93.4} & 44.0 / 11.4\\
\bottomrule
\end{tabular}
\end{table}

We evaluate filter-free safety against CBF-RL (Sec.~\ref{sec:main}), zero-shot scaling and its cost against the joint QP (Sec.~\ref{sec:scaling}), the role of each component (Sec.~\ref{sec:ablations}) and transfer to hardware (Sec.~\ref{sec:hardware}).

\paragraph{Setup.} Agents of radius $0.15$\,m move in a $10\times10$\,m arena with 2 or 4 circular obstacles of radius $0.4$--$1.2$\,m, redrawn every episode ($\Delta t=0.05$\,s, $v_{\max}=2$\,m/s, DI $u_{\max}=6$\,m/s$^2$). In the training \emph{circle} layout, agents start on a circle with antipodal goals, so every episode is an $n$-way crossing through the obstacles at the center, a stress test hardening with $n$; in the \emph{random} layout, starts and goals are uniform. Policies use full observation (MLP actor) or local observation (attention actor, $R=2.5$\,m for SI, $3$\,m for DI). We evaluate the final policy \emph{without filter} on 3 seeds of 100 episodes each, reporting \emph{collisions} ($\%$ of episodes with any collision) and \emph{goals reached} ($\%$ of agents reaching their goal). Every training run of our method recorded zero collisions (App.~\ref{app:impl}).

\paragraph{Baselines.} CBF-RL \citep{yang2025} trains with the same filter and adds its barrier-condition reward and intervention penalty, on the multi-agent SAC of our ablations, so it differs from ``SAC only'' only in the reward (weight screened in App.~\ref{app:baselines}).

\subsection{Filter-Free Safety}\label{sec:main}

With local observation, our method had no collision in any of the $1{,}800$ episodes across both dynamics and all three settings (Table~\ref{tab:main}). CBF-RL matches it with one agent, but with several it collides in $23$--$67\%$ of DI episodes and, with four obstacles, reaches almost no goals, even with its training filter in place (App.~\ref{app:baselines}): its barrier reward penalizes approaching any constraint, and in a crowded crossing every route approaches one. Our full-observation MLP collides noticeably only at $3$ agents and $4$ obstacles ($7$--$8\%$, all with obstacles), cutting inside the margin where two obstacles leave a narrow gap; the local actor waits instead, at a cost of $4$--$6\%$ of goals, a ceiling set by the scenes, where the filter itself reaches only $95\%$. Local sensing is thus the safer observation model.

\subsection{Zero-Shot Scaling under Partial Observability}\label{sec:scaling}

\begin{figure}[t]
\centering
\includegraphics[width=\linewidth]{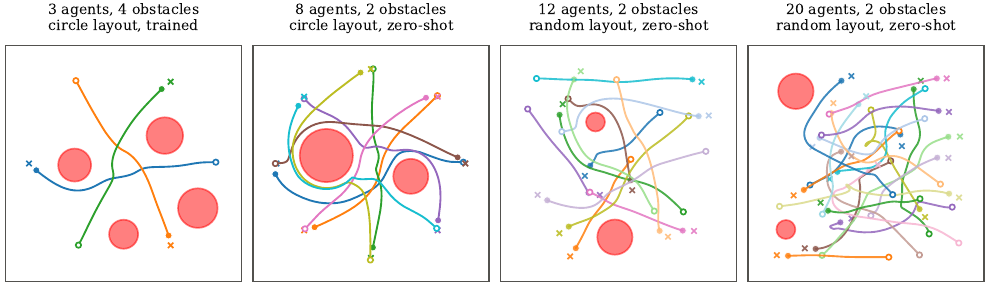}
\caption{Unfiltered roll-outs of one DI policy trained with 3 agents and 4 obstacles (local sensing) in simulation on random selection of layouts tested. Open circles: starts; crosses: goals; filled discs: final positions. The three right panels are zero-shot. All agents reach their goals without collision.}
\label{fig:traj}
\end{figure}

\begin{figure}[t]
\centering
\includegraphics[width=\linewidth]{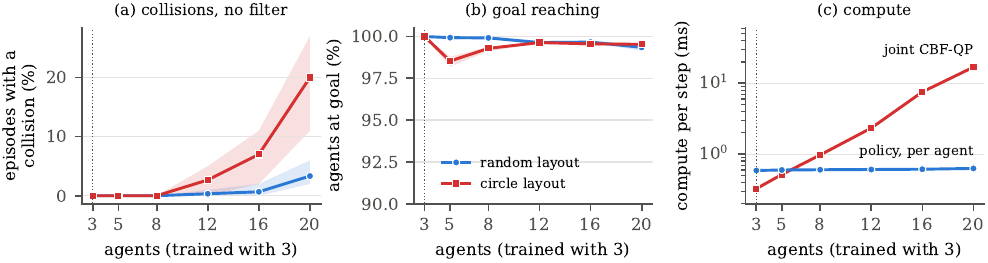}
\caption{Zero-shot scaling of the 3-agent DI policies (local sensing) to up to 20 agents with 2 obstacles, without the filter, in simulation. (a, b) Mean of 3 seeds $\times$ 100 episodes; band: per-seed range. (c) Median per-step compute on one CPU thread: agent's policy versus joint CBF-QP.}
\label{fig:scaling}
\end{figure}

We deploy the three-agent DI policies, without retraining, on teams up to twenty (Figs.~\ref{fig:traj} and~\ref{fig:scaling}). Safety degrades gracefully: up to eight agents, no episode collides in either layout, and in random layouts collisions stay below $1\%$ up to sixteen agents and reach $3.3\%$ at twenty, nearly seven times the training team, reaching over $99\%$ of goals. In the circle stress test, collisions appear only beyond twelve, mostly with obstacles, whose small margin leaves the smallest residual budget (Sec.~\ref{sec:theory}).

\paragraph{Computational cost.} Each agent's local observation policy takes $0.6$\,ms at every team size, without communication (Fig.~\ref{fig:scaling}c), whereas the joint QP grows from $0.3$\,ms at three agents to $16.9$\,ms at twenty, a third of the $50$\,ms control period, before any state is gathered. Removing the filter thus removes the one component whose cost grows with the team.

\subsection{Ablations}\label{sec:ablations}

We remove one component at a time on the hardest setting (DI, local observations, 3 agents and 4 obstacles) and also evaluate each policy zero-shot with 16 agents, where the default policy is not saturated (Table~\ref{tab:ablation}).

\begin{table}[t]
\centering\small
\caption{Component ablation (DI, local sensing $R=3$\,m, trained with 3 agents and 4 obstacles), evaluated without the filter on the training cell and zero-shot with 16 agents. Collisions ($\%$ of episodes) / goals reached ($\%$ of agents), mean over 3 seeds $\times$ 100 episodes in simulation.}
\label{tab:ablation}
\begin{tabular}{lcccc}
\toprule
Variant & RL term & OSFD & 3 agents, 4 obs. & 16 agents, 2 obs.\ (zero-shot)\\
\midrule
\textbf{Ours} & IQL & \checkmark & \textbf{0.0} / \textbf{93.4} & \textbf{7.0} / \textbf{99.6}\\
OSFD only & -- & \checkmark & 0.0 / 93.2 & 2.0 / 92.6\\
IQL only & IQL & -- & 8.0 / 99.9 & 100.0 / 99.9\\
SAC + OSFD & SAC & \checkmark & 0.0 / 95.4 & 13.0 / 99.6\\
SAC only & SAC & -- & 98.3 / 100.0 & 100.0 / 99.7\\
\bottomrule
\end{tabular}
\end{table}

\paragraph{OSFD makes the policy safe beyond the data.} Without OSFD, SAC collides in $98.3\%$ of episodes with the filter removed, the failure predicted by action aliasing, and IQL in $8.0\%$; zero-shot with 16 agents, both collide in every episode, since the buffer contains no crowded near-collision states, the coverage gap of Assumption~\ref{ass:coverage}. With OSFD, no variant collides in training and zero-shot collisions fall to $2$--$13\%$, where in-support extraction with IQL has about half of SAC's ($7.0\%$ vs.\ $13.0\%$).

\paragraph{The RL term drives the policy toward optimality.} OSFD alone imitates the filtered reference controller and receives no reward signal, so it cannot improve on that controller. The RL term supplies this signal: strengthened relative to OSFD, through a larger AWR weight $\beta_d$ or a smaller OSFD weight $w$, it raises goal reaching from $93.2\%$ up to $97.9\%$, while every setting remains collision-free on the training setting (Table~\ref{tab:sweep}, App.~\ref{app:baselines}). Zero-shot collisions grow with it, so $w$ sets how far beyond the training distribution safety extends.

\subsection{Hardware Experiments}\label{sec:hardware}

We fly the 3-agent DI policies on three and five Crazyflie quadrotors \citep{giernacki2017, preiss2017} in a plane at fixed height, with no filter at any point: first the full-observation policies of Table~\ref{tab:main}, with $2$ and $4$ obstacles, then the local-sensing policy, all without retraining. Motion capture provides positions, a ground computer builds each drone's observation and runs the policy, and each drone tracks the training dynamics in a scaled copy of the training arena with stacked-bucket obstacles (App.~\ref{app:hardware}).

With both observation models, every drone reached its goal without contact in $17$ of $18$ flights: $8$ with full observation, $7$ with local sensing and $2$ with five local-sensing drones (Fig.~\ref{fig:hardware}, Table~\ref{tab:hardware} in App.~\ref{app:hardware}). No two drones came closer than $0.18$\,m center to center in these flights (propellers touch at $0.13$\,m). With local sensing, drones start beyond each other's sensing radius and resolve crossing from local information alone.

\section{Conclusion}\label{sec:conclusion}

We present a multi-agent RL framework that trains collision-free with a centralized CBF filter and deploys a decentralized policy with no filter, no communication and only local sensing. This removes the centralized solve whose cost grows with the number of agents. In-support extraction with IQL makes the policy agree with the filter on the data, and off-support safety-filter distillation extends this agreement to near-collision states that the filtered data never contain. The distillation loss provably bounds a closed-form CBF residual of the unfiltered policy, which bounds the barrier uniformly in time and, under coverage, the collision probability at every step. Without a filter, our local-sensing policies had no collisions on their training settings, transferred zero-shot from 3 to 20 agents at constant per-agent cost and flew up to five Crazyflie quadrotors. Since each agent can evaluate its share of every residual locally, future work will monitor the residuals online to decide where to distill next, and will replace motion capture with on-board sensing.
 
\section*{Generative AI Use}

Generative AI tools were used to search the literature, to implement code and experiments, to assist with some mathematical derivations and proofs and to write and edit the paper. The research ideas and methodology are the authors' own. The authors reviewed and verified all AI-assisted content and take responsibility for the final content of the paper.

\section*{Acknowledgments}

This work was supported by the NSF Safe Learning Enabled Systems project, the DARPA TIAMAT project, the U.S. Army Research Laboratory and the U.S. Army Research Office under Grant W911NF2010219, the Office of Naval Research and the National Science Foundation. It used Jetstream2 at Indiana University through allocation CIS251274 from the Advanced Cyberinfrastructure Coordination Ecosystem: Services \& Support (ACCESS) program, which is supported by National Science Foundation grants \#2138259, \#2138286, \#2138307, \#2137603 and \#2138296. We thank Alex Zalles and the HSL group members for reviewing project ideas and manuscript revisions. We also thank Ebonye Smith, Ian Chuang and JoonHo Lee for their help with the hardware experiments.

\bibliographystyle{plainnat}
\bibliography{references}

\newpage
\appendix
 
% \section{Extended Related Work}
% \label{app:related}

% Talk about safety filters; learning-based multi-agent safety; imitation and offline RL.

\section{Safety Filter: Coefficients, Feasibility and Training-Time Safety}
\label{app:filter}

% Coefficients $a_c$ and $b_c$. Convexity of $h_c$. SI feasibility and training being collision-free. DI feasibility.

\paragraph{Coefficients.} Let $\mathcal{K}(x)=\{u\in\mathcal{U}:a_c(x)^T u\ge b_c(x)\ \forall c\}$ denote the admissible CBF action set. Each constraint gives one row $a_c(x)^T u\ge b_c(x)$ of \eqref{eq:filter}, with $a_c=\nabla h_c$ under both dynamics. For SI dynamics, $b_c=-\alpha h_c$. For DI dynamics, $\psi_{1,c}=\nabla h_c^T v+\alpha_1 h_c$ and $\dot\psi_{1,c}+\alpha_2\psi_{1,c}\ge0$ give
\[
b_c=-\big[v^T\nabla^2h_c\,v+(\alpha_1+\alpha_2)\nabla h_c^T v+\alpha_1\alpha_2h_c\big],
\]
where $v^T\nabla^2h\,v=(\|v\|^2-(n^T v)^2)/d$ for a distance barrier at distance $d$ (center to center distance) with unit normal $n$ (relative velocity for pairs), and $0$ for walls. The DI filter also carries hard rows $u_{i,k}\le(v_{\max}-v_{i,k})/\Delta t$ and $u_{i,k}\ge-(v_{\max}+v_{i,k})/\Delta t$, which keep the velocity bound inactive, and it pins the action of agents held at their goals to zero.

\begin{lemma}[Convexity]
\label{lem:convex}
Each barrier in \eqref{eq:barriers} is convex in the joint position $p$, so $h_c(p_w)\ge h_c(p)+\nabla h_c(p)^T w$ for every displacement $w$, with equality for walls.
\end{lemma}
\begin{proof}
    Obstacles and pair barriers are a Euclidean norm of an affine function of $p$ minus a constant; wall barriers are affine.
\end{proof}

\begin{proposition}[SI: feasibility and collision-free training]
\label{prop:si-train}
    For SI dynamics, $0\in\mathcal{K}(x)$ for every $x\in\mathcal{S}$, so the hard filter ($\sigma=0$) is feasible on $\mathcal{S}$. If $x_0\in\mathcal{S}$ and $\alpha\Delta t\le1$, the filtered loop satisfies $h_c(x_t)\ge(1-\alpha\Delta t)^t h_c(x_0)\ge 0$ for every $c$ and $t$: training never collides and the replay buffer contains no collision.
\end{proposition}
\begin{proof}
    At $u=0$ each row reads $0\ge-\alpha h_c(x)$, which holds on $\mathcal{S}$. For the executed action $u_t\in\mathcal{K}(x_t)$, Lemma~\ref{lem:convex} with $w=\Delta t\,u_t$ gives $h_c(x_{t+1})\ge h_c(x_t)+\Delta t\,\nabla h_c^T u_t\ge(1-\alpha\Delta t)\,h_c(x_t)$; induct on $t$.
\end{proof}

\paragraph{DI feasibility.} For DI dynamics, $\{\psi_1\ge0\}\subsetneq\{h\ge0\}$ and states with $\psi_1<0$ are reachable from $\mathcal{S}$, for example by coasting toward an obstacle. There the HOCBF row may be unattainable within $\mathcal{U}$, so the filter uses the relaxing slack ($\sigma=1$): every $u\in\mathcal{U}$ is then feasible, and at an active row the executed action violates the row by at most $s_c=\lambda_c/\kappa$ (from the KKT condition of the slack, with $\lambda_c$ the row's multiplier), which is small at $\kappa=10^4$. This slack bounds the label violation $\epsilon_c$ of Sec.~\ref{sec:theory}. On a radial approach along $\psi_1=0$, the row is attainable without slack whenever $\alpha_1 v_{\max} \le u_{\max}$ ($4\le6$ at our gains). We claim no discrete-time training guarantee for DI dynamics; every DI replay buffer recorded zero collisions.

\section{Learning Objectives in Full}
\label{app:learning}

% Critics and actor in complete form. Also include the entropy term.

The value and critic losses are \eqref{eq:iql}, with twin critics ($k\in\{1,2\}$), the bootstrap term multiplied by $1-d$ for a termination flag $d$, and one output head per agent, since the game is general-sum.

\paragraph{Advantage weights.} In \eqref{eq:awr}, $A_i=\min_k Q^i_{\bar\phi_k}(o,u^{\mathrm{safe}})-V^i_\psi(o)$ and 
\[
W_i=\frac{\tilde W_i}{\mathbb{E}_{\mathcal{B}}[\tilde W_i]},\qquad
\tilde W_i=\big(1+\beta_c\,\mathrm{corr}\big)\min\!\big(\exp(\beta\,\mathrm{clip}(A_i,\pm C)),\,100\big),\qquad
\mathrm{corr}=\min\!\Big(\frac{\|u^{\mathrm{safe}}-u^{\mathrm{raw}}\|}{u_{\max}},1\Big).
\]
The factor $\beta_c=5$ up-weights the transitions in which the filter intervened. Every target remains an executed safe action, so the weights change which safe actions the policy prefers, not whether its targets are safe.

% \begin{proposition}[Feasible-target regression]
% \label{prop:insupport}
%     Fix agent $i$'s observation $o_i$ and a row $a^T u_i\ge b$ whose coefficients are determined by $o_i$. If every regression target paired with $o_i$ satisfies the row, then so does the population minimizer of any positively weighted squared regression onto these targets.
% \end{proposition}

\paragraph{Constraint-violation penalty.}
At the OSFD query states,
\[
\mathcal{L}_{\mathrm{CBF}}=\mathbb{E}_{x\sim\mathcal{D}_q}\Big[\tfrac1n\textstyle\sum_i\sum_{c\ni i}\big(\hat\nu^i_c/u_{\max}\big)^2\Big],
\]
where $\hat\nu^i_c$ is the violation, by agent $i$'s own action $\pi_\theta(o_i)$, of its share of row $c$: a row at least as strict as the filter's for obstacles and walls, and for pairs the reciprocal half $n_{ij}^T u_i\ge n_{ij}^T\tfrac12(y_i+y_j)-\tfrac12\alpha h_{ij}$, an antisymmetric shift of the halves in \eqref{eq:split} that the joint label satisfies whenever it satisfies the pair row. Under DI dynamics the HOCBF rows are used, with each target capped at $u_{\max}\|n_c\|_1$ so that it stays attainable. The penalty only sharpens the policy at the constraint boundary. The analysis uses the exact rows and bounds the residual through $\mathcal{L}_{\mathrm{OSFD}}$.

\paragraph{Full actor objective.} The actor minimizes $\beta_d\mathcal{L}_{\mathrm{AWR}}+w(\mathcal{L}_{\mathrm{OSFD}}+w_\nu\mathcal{L}_{\mathrm{CBF}})+\alpha_{\mathrm{ent}}\mathbb{E}[\log\pi_\theta(\tilde u\mid o_i)]$, where $\tilde u$ is a sampled action and $\alpha_{\mathrm{ent}}$ is tuned toward a target entropy as in SAC \citep{haarnoja2018}. The entropy term only keeps exploration alive. The regression terms act on the mean $\tanh\mu_\theta$, which is the deployed action.

\section{Why IQL over CQL?}
\label{app:iql}

% Next action safety filtering issue during training.

CQL \citep{kumar2020} penalizes $Q$ at actions sampled from the policy, so it evaluates the critic at exactly the raw actions whose values action aliasing leaves unidentified. Its Bellman target must also follow the filtered dynamics: the executed policy is $\Pi_x\circ\pi_\theta$, so every sampled next action must pass through the joint filter. That is one QP per transition, $8{,}192$ per environment step at our batch size and update ratio, or about $3$\,s at $0.4$\,ms per solve, a cost that dominates training and grows with the team. IQL samples no new action: it bootstraps from $V_\psi(o')$ and reads only $(o,u^{\mathrm{safe}},r,o')$ from the buffer, so value learning never calls the filter. The same cost arises for every method whose Bellman target uses a next action sampled from the policy, such as the policy-constraint methods \citep{fujimoto2019, kumar2019bear, fujimoto2021td3bc}; in-sample methods \citep{kostrikov2022, xu2023ivr, garg2023xql} avoid it.

\section{Proofs and local observation certificate for Section~\ref{sec:theory}}
\label{app:proofs}

% Complete proofs from section 4 and indicate local observation directions.

We analyze the deployed loop without collision handling, i.e., loop up to the first collision. Agents held at their goals execute zero action, and so do their labels, so they contribute to neither side of Proposition~\ref{prop:lossres}. For pairs, $\|a_c\|=\sqrt{2}$ because the pair barrier has gradient $(n_{ij},-n_{ij})$ in the joint position.

\begin{proof}[Proof of Theorem~\ref{thm:cert}]
    Each barrier in \eqref{eq:barriers} is convex in the positions, so $h_c(p+w)\ge h_c(p)+\nabla h_c(p)^T w$ for every displacement $w$ (Lemma~\ref{lem:convex}). The SI step is $w=\Delta t \pi(x_t)$, and \eqref{eq:residual} gives $\nabla h_c^T\pi(x_t)\ge-\alpha h_c(x_t)-\nu_c(x_t)$. Hence
    \[
    h_c(x_{t+1})\ \ge\ h_c(x_t)+\Delta t\,\nabla h_c(x_t)^T\pi(x_t)\ \ge\ (1-a)\,h_c(x_t)-\Delta t\,\nu_c(x_t),
    \]
    and unrolling gives \eqref{eq:recursion}. For (i), the geometric sum is at most $\Delta t\,\bar\nu/a=\bar\nu/\alpha$. For (ii), it is less than $\Delta t\,\alpha\rho_c/a=\rho_c$, so $h_c(x_t)>-\rho_c$, whereas a collision requires $h_c<-\rho_c$.
\end{proof}

\begin{proof}[Proof of Corollary~\ref{cor:risk}]
    Fix $c$, let $a=\alpha\Delta t$ and $S_t=\Delta t \sum_{k=0}^{t-1}(1-a)^k \nu_c(x_{t-1-k})$. By Theorem~\ref{thm:cert} and $h_c(x_0)\ge0,$ $\{h_c(x_t)<-\rho_c\}\subseteq\{S_t>\rho_c\}$. Assumption~\ref{ass:coverage} applied to $\nu_c^2\ge0$ gives $\mathbb{E}_{d_t}[\nu_c^2]\le C\mathbb{E}_{\mathcal{D}_q}[\nu_c^2]$ for every step $s<T$, so by Minkowski's inequality
    \[
        \big(\mathbb{E}[S_t^2]\big)^{1/2}\ \le\ \Delta t\sum_{k\ge0}(1-a)^k\big(C\,\mathbb{E}_{\mathcal{D}_q}[\nu_c^2]\big)^{1/2}\ =\ \tfrac{1}{\alpha}\big(C\,\mathbb{E}_{\mathcal{D}_q}[\nu_c^2]\big)^{1/2}.
    \]
    Markov's inequality gives $\Pr(S_t>\rho_c)\le\mathbb{E}[S_t^2]/\rho_c^2$, the first bound of \eqref{eq:prob}; Proposition~\ref{prop:lossres} gives the second, and a union bound over $t\le T$ the episode-level claim. The proof uses only the moment condition $\mathbb{E}_{d_t}[\nu_c^2]\le C\mathbb{E}_{\mathcal{D}_q}[\nu_c^2]$, which we measure in App.~\ref{app:audit}.
\end{proof}

\begin{lemma}[The train-deploy shift arises only through the residual]\label{lem:noshift}
    For the hard SI filter, if $\nu_c(x_t)=0$ for every c along the filtered trajectory from $x_0$, then the unfiltered trajectory from $x_0$ coincides with it.
\end{lemma}
\begin{proof}
    If $\pi(x_t)\in\mathcal{K}(x_t)$, then $(u,s)=(\pi(x_t),0)$ attains the minimum $0$ of \eqref{eq:filter}, so $\Pi_{x_t}(\pi(x_t))=\pi(x_t)$; induct on $t$.
\end{proof}

\paragraph{Local observation.} The following results hold for SI dynamics.
\begin{proposition}[The pair label depends on the neighbour's intent]\label{prop:labelvar}
    With only the pair row $(i,j)$ active and the box inactive, the filter's output for agent $i$ is $y_i=r_i-\tfrac12(\alpha h_{ij}+n_{ij}^T r_i-n_{ij}^T r_j)\,n_{ij}$, where $r_i,r_j$ are the actions passed to the filter. If $o_i$ does not determine $n_{ij}^T r_j$, every decentralized policy has $\mathbb{E}[\|\pi_\theta(o_i)-y_i\|^2\mid o_i]\ge\tfrac14\mathrm{Var}(n_{ij}^T r_j\mid o_i)$ when the row is active.
\end{proposition}
\begin{proof}
    Projecting onto the active half-space, with $a=(n_{ij},-n_{ij})$ and $\|a\|^2=2$, moves $r$ by $\lambda a$ with $\lambda=[-\alpha h_{ij}-a^T r]_+/2$. Given $o_i$, the only random term in $y_i$ is $\tfrac12(n_{ij}^T r_j)\,n_{ij}$, and the best squared-loss predictor is its conditional mean, whose error is the conditional variance.
\end{proof}

Safety does not require matching the label. A pair row splits into half per agent,
\begin{equation}
\label{eq:split}
n_{ij}^T(u_i-u_j)+\alpha h_{ij}
=\underbrace{\big(n_{ij}^T u_i+\tfrac12\alpha h_{ij}\big)}_{\text{agent } i}
+\underbrace{\big(n_{ji}^T u_j+\tfrac12\alpha h_{ij}\big)}_{\text{agent } j},\qquad n_{ji}=-n_{ij},
\end{equation}
and agent $i$'s violation of its half, $\hat{\nu}_{ij}^i=\big[-\big(n_{ij}^T\pi_\theta(o_i)+\tfrac12\alpha h_{ij}\big)\big]_+$ depends only on its own action and its token $\tau_{ij}$.

\begin{proposition}[Decentralized pair certificate]\label{prop:pair}
    For every pair, $\nu_{ij}\le \hat{\nu}_{ij}^i+\hat{\nu}_{ij}^j$. Hence, if every agent keeps $\hat{\nu}_{ij}^i<\frac12\alpha\rho_{\mathrm{pair}}$ for every agent $j$ at every step, no agent-agent collision occurs.
\end{proposition}
\begin{proof}
    With $A,B$ the two halves in \eqref{eq:split}, $\nu_{ij}=[-(A+B)]_+\le[-A]_++[-B]_+<\alpha\rho_{\mathrm{pair}}$, and Theorem~\ref{thm:cert}(ii) applies to the pair barrier.
\end{proof}

The same holds for any antisymmetric shift of the halves, such as the reciprocal split used in training (App.~\ref{app:learning}), and the HOCBF rows of DI dynamics split in the same way.

\paragraph{Unsensed constraints are inactive.} An unsensed agent has $h_{ij}>R-2r-\rho_{\mathrm{pair}}$; so agent $i$'s half-row holds for every admissible action whenever $\frac12\alpha(R-2r-\rho_{\mathrm{pair}})\ge\sqrt2 u_{\max}$; at our SI settings the two sides are $3.0$ and $2.83$\,m/s. An unsensed obstacle's row holds since $\alpha(R-r-\rho)\approx6.8\ge2.83\,$m/s, and walls are always observed. No constraint that the policy could violate lies outside the sensing radius.

\paragraph{Obstacles and walls, agent by agent.} For any obstacle or wall row of agent $i$, $a_c$ acts only on agent $i$'s action, has unit norm, and is determined by $o_i$ when the obstacle is sensed. The proofs above then go through with $\nu_c\le\|\pi_\theta(o_i)-y_i(x)\|+\epsilon_c$ and with Assumption~\ref{ass:coverage} stated for the distribution of agent $i$'s observation rather than the joint state.
 
\section{Double Integrator Certificate}
\label{app:di}

% Double integrator adoption with HOCBF.
The DI labels satisfy each HOCBF row up to the slack, so $\epsilon_c\le s_c$ and Proposition~\ref{prop:lossres} applies unchanged. For the barrier, write $a_i=\alpha_i\Delta t$; for a distance constraint let $w$ be the relative velocity and $d_c$ the contact distance ($r_o+r$ or $2r$). We assume, up to step $T$, that the velocity bound is inactive and that $\|w_k\|\le V_c$ with $\Delta t V_c<d_c$. Before the first collision, every segment between samples then stays at distance at least $d_{*,c}=d_c-\Delta tV_c$ from the obstacle center (or the other agent), since each sample is at distance at least $d_c$ and each segment has length at most $\Delta t\,V_c$. Along such segments, $\|\nabla^2 h\|\le1/d_{*,c}$ and $\nabla^2 h$ is Lipschitz with constant $L_{H,c}=2/(\sqrt3 d_{*,c}^2)$ (Lemma~\ref{lem:curv}); for walls both are zero. These are hypotheses on the deployed trajectory, not consequence of training.

\begin{lemma}[Curvature of a distance barrier]
\label{lem:curv}
Let $h(p)=\|p-c_o\|-\bar d$ on $\mathbb{R}^2$ and, at a point with $d=\|p-c_o\|>0$, let $n=(p-c_o)/d$. Then $\nabla^2h(p)=(I-nn^T)/d$, so $0\preceq\nabla^2h(p)\preceq I/d$ and $\|\nabla^2h(p)\|=1/d$. If the segment from $q$ to $p$ stays at distance at least $d_*$ from $c_o$, then $|v^T(\nabla^2h(p)-\nabla^2h(q))\,v|\le L_H\|p-q\|\,\|v\|^2$ for every $v$, with $L_H=2/(\sqrt3\,d_*^2)$.
\end{lemma}
\begin{proof}
Since $\nabla h=n$ and $\nabla n=(I-nn^T)/d$, the Hessian is $(I-nn^T)/d$, with eigenvalue $0$ along $n$ and $1/d$ along the tangent $n^\perp$. For the Lipschitz bound, fix a unit $v$ and differentiate $g(p)=v^T\nabla^2h(p)\,v=(1-(n^T v)^2)/d$ along a unit direction $u$, using $\partial_un=(u-(n^T u)\,n)/d$ and $\partial_ud=n^T u$:
\[
\partial_ug=-\frac{2(n^T v)\big(u^T v-(n^T u)(n^T v)\big)+(n^T u)\big(1-(n^T v)^2\big)}{d^2}.
\]
Writing $v=\cos\vartheta\,n+\sin\vartheta\,n^\perp$, this is $\partial_ug=-\big(\sin^2\!\vartheta\,(n^T u)+\sin2\vartheta\,(n^{\perp T}u)\big)/d^2$, whose largest magnitude over unit $u$ is $\sqrt{\sin^4\vartheta+\sin^22\vartheta}\,/d^2=|\sin\vartheta|\sqrt{4-3\sin^2\vartheta}\,/d^2$. Over $\vartheta$ this peaks at $\sin^2\vartheta=2/3$, where it equals $2/(\sqrt3\,d^2)$. Integrating $\partial_ug$ along the segment from $q$ to $p$, on which $d\ge d_*$, and scaling by $\|v\|^2$ gives the bound.
\end{proof}

\begin{theorem}[Sampled-data DI certificate]\label{thm:di} Let $a_1\in(0,1)$ and $a_2(1+a_1)\in(0,1]$. Start at rest in $\mathcal{S}$, assume the conditions above up to step $T$, and let $\bar{\nu}=\max_{k<T}\nu_c(x_k)$ for the HOCBF row of constraint $c$. Then for all $t\le T$,
\begin{equation}
\label{eq:di}
h_c(x_t)\ \ge\ -\frac{\bar\nu}{\alpha_1\alpha_2(1+\alpha_1\Delta t)}\ -\ \Gamma_c,
\end{equation}
with
\[
\Gamma_c=\frac{\bar\gamma_c}{\alpha_1\alpha_2\Delta t\,(1+a_1)},\qquad
\bar\gamma_c=\Delta t^2\Big(\tfrac{L_{H,c}}{2}V_c^3+\tfrac{\alpha_1+\alpha_2}{d_{*,c}}V_c^2+\alpha_1^2\alpha_2\,\rho_c\Big),
\]
so $\Gamma_c=O(\Delta t)$. In particular, if $\Gamma_c<\rho_c$ and $\bar\nu<\alpha_1\alpha_2(1+a_1)(\rho_c-\Gamma_c)$, no collision on $c$ occurs before step $T$.
\end{theorem}
\begin{proof}
    \emph{Position step.} Lemma~\ref{lem:convex} with $w=\Delta t\,v_{k+1}$ gives $h_{k+1}\ge(1-a_1)h_k+\Delta t\,\psi_k$, with the discrete HOCBF function $\psi_k=n_k^T v_{k+1}+\alpha_1h_k$.
\emph{Velocity step.} Let $\eta=n_{k+1}^T v_{k+1}$ and $c=v_{k+1}^T\nabla^2h(p_{k+1})\,v_{k+1}$. The residual's definition at $x_{k+1}$ gives $n_{k+1}^T u_{k+1}\ge-c-(\alpha_1+\alpha_2)\eta-\alpha_1\alpha_2h_{k+1}-\nu_{k+1}$, so
\[
\psi_{k+1}=\eta+\Delta t\,n_{k+1}^T u_{k+1}+\alpha_1h_{k+1}\ \ge\ (1-a_1-a_2)\,\eta-\Delta t\,c-\Delta t\,\nu_{k+1}+\alpha_1(1-a_2)\,h_{k+1}.
\]
Since $n_{k+1}-n_k=\Delta t\,M_kv_{k+1}$ with $M_k=\int_0^1\nabla^2h(p_k+s\Delta t\,v_{k+1})\,ds\succeq0$, we have $\eta=\psi_k-\alpha_1h_k+\Delta t\,C_k$ with $C_k=v_{k+1}^T M_kv_{k+1}$. Substituting this and the position step,
\[
\psi_{k+1}\ \ge\ \lambda\,\psi_k-\Delta t\,\nu_{k+1}-\gamma_k,\quad \lambda=1-a_2(1+a_1)\in[0,1),\quad
\gamma_k=\Delta t\big(c-(1-a_1-a_2)C_k\big)-\alpha_1a_1a_2\,h_k\ \le\ \bar\gamma_c,
\]
using $|c-C_k|\le\tfrac{L_{H,c}}{2}\Delta t\,V_c^3$ (Lemma~\ref{lem:curv} with $p_{k+1}-(p_k+s\Delta t\,v_{k+1})=(1-s)\Delta t\,v_{k+1}$ and $\int_0^1(1-s)\,ds=\tfrac12$), $0\le C_k\le V_c^2/d_{*,c}$ (Lemma~\ref{lem:curv}) and $h_k\ge-\rho_c$ before the first collision.
\emph{Cascade.} At rest, $\psi_0\ge-\Delta t\,\bar\nu$. Unrolling the velocity step, $\psi_k\ge-(\Delta t\,\bar\nu+\bar\gamma_c)/(a_2(1+a_1))$; unrolling the position step from $h_0\ge0$ gives $h_k\ge-(\Delta t\,\bar\nu+\bar\gamma_c)/(\alpha_1a_2(1+a_1))$, which is \eqref{eq:di}.
\end{proof}

For a pair, the same argument applies in relative positions and velocities: in the joint position, $v^T\nabla^2h_{ij}\,v=w^T H\,w$ with $w=v_i-v_j$ and $H$ the Hessian of Lemma~\ref{lem:curv} at $p_i-p_j$. The bound is conservative, but not by much at our control step: in simulation, an adversarial tangential approach at $\Delta t=0.05\,$s, $(\alpha_1,\alpha_2)=(2,4)$ and $\rho=0.1\,$m produces clearance deficits of $0.08-0.10\,$m, nearly the whole margin, while radial approaches produce none. This is why we measure DI safety empirically.

\section{Audit of the Bounds}
\label{app:audit}

% Table with different values noted for different methods to show correlation.
\paragraph{Protocol.} For each trained policy we rebuild $\mathcal{D}_q$ at its training cell as in training, label it with the training filter, and run $100$ unfiltered episodes of the deterministic policy, recording the label error $e_\theta$, the residual $\nu_c$ and the barrier $h_c$ of every row at every step. We audit 75 policies: 21 SI policies (ours with full and local observation, CBF-RL and SAC) and the 54 DI policies of Tables~\ref{tab:ablation} and \ref{tab:sweep} and of the 2-obstacle setting, at the training cell and, for the $51$ ablation policies, zero-shot with $16$ agents.

\begin{table}[h]
\centering\small
\caption{Audit of the chain \eqref{eq:chain} without the filter. On $\mathcal{D}_q$ at the training cell: the OSFD loss and the root-mean-square label error $e_\theta$ and residual $\nu$; on the deployed states: the same error and residual, and collisions over 100 episodes per seed. $\nu$ is the larger of the obstacle/wall and pair groups; units m/s (SI) and m/s$^2$ (DI). Means over 3 seeds; $^\dagger$one seed.}
\label{tab:audit}
\begin{tabular}{llccccccc}
\toprule
& & & \multicolumn{3}{c}{On $\mathcal{D}_q$} & \multicolumn{3}{c}{Deployed}\\
\cmidrule(lr){4-6}\cmidrule(lr){7-9}
& Policy & Agents/obs. & $\mathcal{L}_{\mathrm{OSFD}}$ & $e_\theta$ & $\nu$ & $e_\theta$ & $\nu$ & Coll.\ (\%)\\
\midrule
SI & Ours, local & 5/4 & 0.012 & 0.49 & 0.05 & 0.40 & 0.05 & 0.7\\
SI & Ours, local, zero-shot & 16/2 & 0.012 & 0.49 & 0.05 & 1.70 & 0.17 & 16.3\\
SI & Ours, full obs. & 5/4 & 0.060 & 1.09 & 0.10 & 1.00 & 0.05 & 4.3\\
SI & Ours, full obs. & 8/4 & 0.083 & 1.63 & 0.13 & 1.46 & 0.13 & 10.7\\
SI & Ours, full obs.$^\dagger$ & 3/8 & 0.327 & 1.98 & 0.38 & 1.95 & 0.26 & 98.0\\
SI & SAC only$^\dagger$ & 3/4 & 0.938 & 3.36 & 0.78 & 3.47 & 0.13 & 100.0\\
SI & CBF-RL ($w=3$)$^\dagger$ & 3/4 & 1.459 & 4.18 & 0.81 & 3.39 & 0.04 & 22.0\\
\midrule
DI & Ours, local & 3/4 & 0.032 & 1.78 & 1.16 & 1.12 & 0.06 & 0.0\\
DI & Ours, local, zero-shot & 16/2 & 0.032 & 1.78 & 1.16 & 5.09 & 0.50 & 7.7\\
DI & OSFD only, zero-shot & 16/2 & 0.026 & 1.63 & 1.18 & 5.29 & 0.29 & 0.7\\
DI & IQL only & 3/4 & 0.550 & 7.38 & 4.43 & 5.63 & 1.98 & 8.3\\
DI & SAC only & 3/4 & 1.802 & 13.45 & 9.07 & 15.48 & 10.19 & 97.7\\
\bottomrule
\end{tabular}
\end{table}

\paragraph{Each link of the chain holds.} The three links of \eqref{eq:chain} were checked directly. \emph{Loss to label error:} on $\mathcal{D}_q$, $\mathbb{E}[e_\theta^2]$ equals $n\,u_{\max}^2\mathcal{L}_{\mathrm{OSFD}}$ for SI policies and is $87-97\%$ of it for DI policies, whose goal-held agents are excluded from the loss. \emph{Label error to residual:} Proposition~\ref{prop:lossres} held on all $6.8\times 10^7$ rows evaluated. \emph{Residual to barrier:} along the unfiltered SI roll-outs, the one-step inequality behind \eqref{eq:recursion}
held on all $2.8\times 10^7$ row-steps before the first collision, and each of the 724 SI collisions, $503$ of them by SAC and CBF-RL, was preceded by a discounted residual sum above $\rho_c$, as Theorem~\ref{thm:cert} requires.

\paragraph{Lowering one quantity lowers the next.} Across the $21$ policies, Spearman rank correlations along the chain are $0.97$ from $\mathcal{L}_{\mathrm{OSFD}}$ to $e_\theta$, $0.98$ from $e_\theta$ to $\nu$ on $\mathcal{D}_q$, and $0.88$ from $\nu$ to the collision rate; the loss alone predicts collisions at $0.86$ (Table~\ref{tab:audit}, top). Under DI dynamics, $\mathcal{L}_{\mathrm{OSFD}}$ and $e_\theta$ on $\mathcal{D}_q$ are ranked identically, and on the deployed states $e_\theta$ predicts $\nu$ at $0.90$. On $\mathcal{D}_q$ itself, the DI residual of every policy trained with OSFD is $1.15$--$1.25$\,m/s$^2$ whatever its loss: the synthetic states include approaches that no admissible acceleration can stop, where the label's own violation $\epsilon_c$ dominates $\nu_c$. The two variants trained without OSFD have losses of $0.55$ and $1.8$, against at most $0.063$ with OSFD, residuals on $\mathcal{D}_q$ of $4.4$ and $9.1$\,m/s$^2$, and are the only DI policies that collide on the training scenes. 

% Zero-shot with 16 agents, the loss rank-correlates with collisions at $0.79$ across the 51 DI ablation policies.

\paragraph{Coverage.} The proof Corollary~\ref{cor:risk} needs only $\mathbb{E}_{d_t}[\nu_c^2]\le C \mathbb{E}_{\mathcal{D}_q}[\nu_c^2]$, which we measure as the smallest such $C$ over 50-step windows. On the training scenes it is at most $0.83$ for the obstacle and wall rows of every SI policy of our method and $0.04$ for every DI policy trained with OSFD, since $\mathcal{D}_q$ concentrates on near-collision states; pair rows need up to $8$ (SI) and $0.18$ (DI). 

% Zero-shot with $16$ agents, the SI local policy needs $4-6$ on obstacle and wall rows, and that is where its collisions appear: the crowded states lie outside $\mathcal{D}_q$.

\section{Implementation Details}
\label{app:impl} 

\begin{table}[t]
\centering\small
\caption{Hyperparameters.}
\label{tab:hyper}
\begin{tabular}{lll}
\toprule
Group & Single integrator & Double integrator\\
\midrule
Environment & $\Delta t=0.05$\,s, $v_{\max}=2$\,m/s, $r=0.15$\,m, 1000 steps & same, $u_{\max}=6$\,m/s$^2$\\
Filter & $\alpha=3$, $\rho=0.075$, $\rho_{\mathrm{pair}}=0.2$ & $\alpha_1=2$, $\alpha_2=4$, $\kappa=10^4$, $\rho=0.1$\\
IQL & \multicolumn{2}{l}{$\tau=0.7$, $\beta=3$, weight cap 100, $\beta_d=10$, $\beta_c=5$, target entropy $-0.2\,\dim u_i$}\\
OSFD & $w=50$, $w_\nu=10$, 256 states, mixture (0.4, 0.3, 0.3) & same, adverse speed $\mathcal{U}[0,1]v_{\max}$\\
Sensing & $R=2.5$\,m & $R=3.0$\,m\\
Optimization & \multicolumn{2}{l}{Adam, lr $6\times10^{-4}$, batch 4096, $\gamma=0.99$, Polyak 0.005, 2 updates/step, 16 envs}\\
Exploration & \multicolumn{2}{l}{5k goal-directed warm-up steps (noise 0.5); goal-directed proposals w.p.\ 0.3 per env}\\
Training & \multicolumn{2}{l}{200k steps (1 agent), 300k steps (3 agents)}\\
\bottomrule
\end{tabular}
\end{table}

% Networks used and their sizes. \red{We should include some training curves.}
\paragraph{Scenes and reward.} The arena is $[-5,5]^2$ ($L=5\,$m), $r=0.15\,$m, and obstacle radii lie in $[0.4,1.2]\,$m, redrawn every episode. Circle layout: starts evenly spaced on a circle of radius $4$\,m with antipodal goals. Random layout: starts and goals uniform in $[-4,4]^2$, pairwise at least $1$\,m apart, each start at least $4$\,m from its goal. An agent within $0.5$\,m of its goal is held at rest. The reward is $r_i=\gamma\Phi(p_i')-\Phi(p_i)+0.02\langle\hat u_i,\hat g_i\rangle+300(1+\tfrac12(1-t/T))\mathbf{1}[\text{reach}]-30\,\mathbf{1}[\text{obstacle}]-10\,\mathbf{1}[\text{wall}]-30\,\mathbf{1}[\text{agent}]-20\,\mathbf{1}[\text{timeout}]$, with $\Phi=-\|p_i-g_i\|$ \citep{ng1999}; under DI dynamics the directional term uses the velocity, and under the filter the collision terms never fire.

\paragraph{Observations and networks.} Full observation: own position, goal offset, offsets to all agents and obstacles, obstacle radii, wall clearances, previous action and, under DI dynamics, all velocities divided by $v_{\max}$. Local observation: the tokens of Sec.~\ref{sec:po}, 8-dimensional, whose motion slot holds the relative velocity divided by $v_{\max}$ (DI) or the neighbor's last action (SI); masking by $R$ is applied before any learned layer. Actor: squashed Gaussian, a $2\times256$ MLP with LayerNorm (full observation) or a per-token MLP of width 128 and 4-head attention followed by a $2\times256$ trunk (local observation). Twin critics and value network: $2\times512$ MLPs with LayerNorm \citep{ba2016} over all agents' observations, one head per agent. All networks use orthogonal initialization and Adam \citep{kingma2015}.

\paragraph{Training loop and OSFD.} Sixteen environments run in parallel. For the first 5k steps, proposals are goal-directed with Gaussian noise; afterwards each agent samples $u^{\mathrm{raw}}_i\sim\pi_\theta(\cdot\mid o_i)$, replaced by the goal-directed proposal with probability $0.3$ per environment to keep goal-reaching transitions in the buffer. Each environment step is followed by two updates, each sampling 4096 transitions and building 256 OSFD query states. $\mathcal{D}_q$ mixes synthetic boundary states ($40\%$), buffer states ($30\%$) and buffer states with positions perturbed by $\mathcal{N}(0,0.3^2)$ and DI velocities by $\mathcal{N}(0,0.5^2)$ ($30\%$). A synthetic state keeps a buffer row's obstacles and goals and places agents within $0.6$\,m of an obstacle, a wall or one another, heading toward it (under DI dynamics at speed $\mathcal{U}[0,1]\,v_{\max}$). The reference is $u^{\mathrm{ref}}_i=u_{\max}w_i/\|w_i\|$ with $w_i=\hat g_i+\sum_k[1-\Delta_{ik}]_+[-\hat g_i^T n_{ik}]_+\,t(n_{ik})$, clearance $\Delta_{ik}$ and clockwise tangent $t(n)=(n_y,-n_x)$; under DI dynamics, $u^{\mathrm{ref}}_i=\mathrm{clip}(3(v_{\max}w_i/\|w_i\|-v_i),\pm u_{\max})$. All QPs are solved with the Goldfarb-Idnani method \citep{goldfarb1983}. The reported policy is the final actor.

\begin{figure}[t]
\centering
\includegraphics[width=\linewidth]{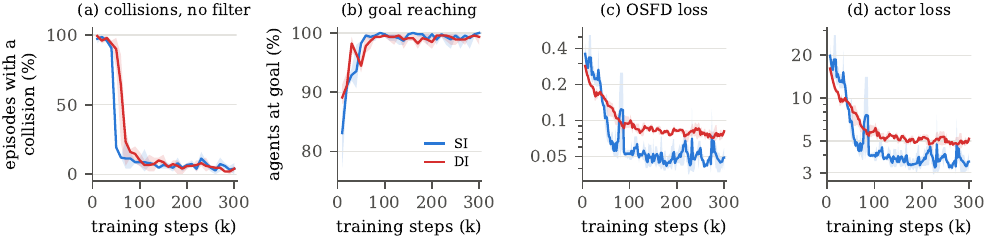}
\caption{Training with full observation, 3 agents and 4 obstacles: SI (blue) and DI (red), mean over 3 seeds; band: per-seed range. (a, b) Filter-free evaluation of the current policy on the training scenes, 50 episodes every 10k steps. (c) The OSFD term $\mathcal{L}_{\mathrm{OSFD}}+w_\nu\mathcal{L}_{\mathrm{CBF}}$ and (d) the full actor objective (App.~\ref{app:learning}), logged every 2k steps on one batch and smoothed by a running median over 10k steps; logarithmic axes.}
\label{fig:training}
\end{figure}

\paragraph{Training curves.} Figure~\ref{fig:training} shows the training of the full-observation policies of Table~\ref{tab:main} with $3$ agents and $4$ obstacles. Every $10$k steps, the current policy is evaluated without the filter on $50$ episodes of the training scenes. The filtered training rollouts never collide, yet the unfiltered policy initially collides in almost every episode. As the OSFD loss falls, filter-free collisions drop below $20\%$ within $50$k (SI) and $80$k (DI) steps and stay near $5\%$ over the last $100$k steps, while goal reaching rises above $99\%$ (final values in Table~\ref{tab:main}). The actor loss follows the OSFD term, which makes up most of it at $w=50$.

\section{Baselines and Hyperparameters}
\label{app:baselines}

\begin{table}[t]
\centering\small
\caption{One-hyperparameter sweeps around the default ($^\star$) on the cell of Table~\ref{tab:ablation}, unfiltered. Collisions / goals (\%), mean over 3 seeds.}
\label{tab:sweep}
\begin{tabular}{llcc}
\toprule
Hyperparameter & Value & 3 agents, 4 obs. & 16 agents, 2 obs.\ (zero-shot)\\
\midrule
OSFD weight $w$ & 10 & 0.0 / 97.9 & 36.0 / 99.7\\
 & 25 & 0.0 / 94.7 & 18.3 / 99.7\\
 & 50$^\star$ & 0.0 / 93.4 & 7.0 / 99.6\\
 & 100 & 0.0 / 93.3 & 1.3 / 99.6\\
\midrule
AWR weight $\beta_{\mathrm d}$ & 3 & 0.0 / 93.8 & 1.7 / 99.5\\
 & 10$^\star$ & 0.0 / 93.4 & 7.0 / 99.6\\
 & 30 & 0.0 / 96.4 & 36.0 / 99.6\\
\midrule
Expectile $\tau$ & 0.6 & 0.0 / 93.2 & 4.3 / 99.6\\
 & 0.7$^\star$ & 0.0 / 93.4 & 7.0 / 99.6\\
 & 0.8 & 0.0 / 94.2 & 12.0 / 99.6\\
 & 0.9 & 0.0 / 94.2 & 13.3 / 99.4\\
\midrule
Training radius $R$ (m) & 1.5 & 0.0 / 94.7 & 13.0 / 99.8\\
 & 2.0 & 0.0 / 93.9 & 12.0 / 99.6\\
 & 2.5 & 0.0 / 94.2 & 15.7 / 99.6\\
 & 3.0$^\star$ & 0.0 / 93.4 & 7.0 / 99.6\\
 & $\infty$ (full) & 0.0 / 95.0 & 13.0 / 99.6\\
\bottomrule
\end{tabular}
\end{table}

% CBF-RL extension to MAS and hyperparameters used.
All baselines use the same filter, environment, networks and evaluation. \textbf{CBF-RL} adds to each agent's reward $w\big[\sum_c\min(a_c^T u^{\mathrm{raw}}-b_c,0)/(s_cu_{\max})+\exp(-\|u^{\mathrm{raw}}-u^{\mathrm{safe}}\|^2/\sigma^2)-1\big]$, the barrier term and bounded intervention penalty of \citet{yang2025}, computed with the rows of the joint filter (a pair row is charged to both agents), with $\sigma=0.25u_{\max}$ and $s_c$ the row scale, and trains with multi-agent SAC. A screen over $w\in\{0.3,1,3\}$ (3 agents, 4 obstacles, one seed) gave the fewest collisions at $w=3$ on both dynamics. With its training filter still on, CBF-RL reaches only $0$--$13\%$ of goals with 3 agents and 4 obstacles, so its low goal rate there is not caused by removing the filter. \textbf{SAC only} runs SAC on the raw action, with the filter treated as part of the environment; \textbf{SAC + OSFD} adds the OSFD terms; \textbf{IQL only} sets $w=0$; \textbf{OSFD only} sets $\beta_d=0$, so the critics train but do not reach the actor. Table~\ref{tab:sweep} varies one hyperparameter at a time around the default.

\section{Hardware Setup}
\label{app:hardware}

% Lab setup and runs, if done in time.
\begin{table}[t]
\centering\small
\caption{Hardware flights without a filter (DI, trained with 3 agents). The five-drone flights are zero-shot in the number of drones. Separation: smallest center-to-center distance between two drones. Clearance: smallest horizontal distance from a propeller tip to an obstacle rim (virtual obstacles: to the declared disc). Time: until the last drone reaches its goal.}
\label{tab:hardware}
\begin{tabular}{lcccccc}
\toprule
Policy and layout & Flights & Collisions & Goals (\%) & Separation (m) & Clearance (m) & Time (s)\\
\midrule
\multicolumn{7}{l}{\emph{Local sensing, 3 drones, 4 obstacles}}\\
\quad Scattered & 6 & 0 & 100 & 0.25 & 0.03 & 10.7--13.4\\
\quad Row, spread & 1 & 0 & 100 & 0.19 & 0.23 & 9.4\\
\quad Row, tight & 1 & 1 & 66.7 & contact & -- & --\\
\multicolumn{7}{l}{\emph{Local sensing, 5 drones, 2 obstacles (zero-shot)}}\\
\quad Circle layout & 2 & 0 & 100 & 0.20 & 0.16 & 10.4--11.4\\
\multicolumn{7}{l}{\emph{Full observation, 3 drones}}\\
\quad 4 obstacles & 2 & 0 & 100 & 0.26 & 0.16 & 12.9--13.6\\
\quad 2 obstacles & 3 & 0 & 100 & 0.30 & 0.10 & 10.1--18.2\\
\quad 2 virtual obstacles & 3 & 0 & 100 & 0.18 & 0.04 & 9.5--13.6\\
\bottomrule
\end{tabular}
\end{table}

\paragraph{Platform and scaling.} Three or five Crazyflie 2.x quadrotors \citep{giernacki2017} fly at $0.8$\,m through Crazyswarm2 \citep{preiss2017}, tracked by a Vicon system at 100\,Hz, with a Mellinger controller tracking full-state setpoints. With length scale $s=0.375$, time stretch $k=2$ and arena center $c$, we map $p^{\mathrm{lab}}=c+s\,p$, $v^{\mathrm{lab}}=(s/k)\,v$, $u^{\mathrm{lab}}=(s/k^2)\,u$ and $\Delta t$ to $k\Delta t$, so a trajectory of the training plant is a lab trajectory slowed down by $k$: walls at $\pm1.875$\,m, a 10\,Hz loop and a sensing radius of $1.125$\,m.

\paragraph{Control loop.} Every $0.1$\,s a laptop maps the Vicon positions to the training frame, builds each drone's observation exactly as in training (masked to the sensing radius), evaluates $u_i=u_{\max}\tanh\mu_\theta(o_i)$ in $0.46$\,ms for all three drones ($0.60$\,ms for five) with no QP, integrates it with the training plant from the measured position, and sends the result to the drone as a setpoint, which was tracked to within $2.4$\,cm (95th percentile) in every flight. The velocities the policy sees are the commanded plant velocities. A drone within $0.6$ training units of its goal ($0.5$ in training, enlarged so that a real drone has room to brake) is held in place. The drones exchange nothing; motion capture stands in for on-board sensing.

\paragraph{Obstacles and flights.} Each obstacle is a stack of three buckets of rim radius $0.16$\,m, declared to the policy with radius $0.30$\,m ($0.8$ training units) to absorb the drone's size, its tracking error and the gap between the lab and the training plant; over the 18 flights, the agent's disc entered a declared disc in four, and no propeller came within $3$\,cm of a rim. The five-drone flights use the training policy (3 agents, 4 obstacles) unchanged, with five starts and antipodal goals evenly spaced on the training circle and two obstacles; neighbouring drones start $4.6$ training units apart, beyond the sensing radius of $3$, so the drones sense each other only once the crossing begins. The first full-observation flights used virtual obstacles. Each flight starts with the drones hovering at rest at the starts of the circle layout and ends when all are held at their goals. In the tight row layout, two drones met in the central opening and the policy's setpoints converged until they touched, so the contact came from the policy rather than from tracking error. Videos of the flights are available at \projectsite.

\end{document}